\documentclass{article}
\usepackage{ijcai25}

\usepackage{times}
\usepackage{soul}
\usepackage{url}
\usepackage[hidelinks]{hyperref}
\usepackage[utf8]{inputenc}
\usepackage[small]{caption}
\usepackage{graphicx}
\usepackage{amsmath}
\usepackage{amsthm}
\usepackage{booktabs}
\usepackage{algorithm}
\usepackage{algorithmic}
\usepackage[switch]{lineno}

\usepackage{xcolor}
\usepackage{graphicx}

\title{PromptIQ: Who Cares About Prompts? Let System Handle It – A Component-Aware Framework for T2I Generation}

\author{
Nisan Chhetri$^1$
\and
Arpan Sainju$^2$\\
\affiliations
$^1$North Carolina State University\\
$^2$Middle Tennessee State University \\
\emails
nchhetr@ncsu.edu,
arpan.sainju@mtsu.edu
}

\begin{document}

\maketitle

\begin{abstract}

Generating high-quality images without prompt engineering expertise remains a challenge for text-to-image (T2I) models, which often misinterpret poorly structured prompts, leading to distortions and misalignments. While humans easily recognize these flaws, metrics like CLIP fail to capture structural inconsistencies, exposing a key limitation in current evaluation methods. To address this, we introduce PromptIQ, an automated framework that refines prompts and assesses image quality using our novel Component-Aware Similarity (CAS) metric, which detects and penalizes structural errors. Unlike conventional methods, PromptIQ iteratively generates and evaluates images until the user is satisfied, eliminating trial-and-error prompt tuning. Our results show that PromptIQ significantly improves generation quality and evaluation accuracy, making T2I models more accessible for users with little to no prompt engineering expertise. 
\end{abstract}

\section{Introduction}
The rapid advancements in T2I generation have opened new possibilities across various domains, making it a crucial area of research. For example, T2I can enhance efficiency and accessibility in creative industries, reduce costs in marketing and design, and support education through visual learning tools ~\cite{shelby2024generative,singh2024transforming,montenegro2024integrative}.
Despite its potential, generating high-quality images through T2I models remains a challenge, especially for users with limited prompt engineering knowledge ~\cite{liu2022design,wang2024promptcharm}. Many existing models require precise and well-structured prompts to produce desirable results, often leading to frustration and inefficiency ~\cite{hao2024optimizing,mo2024dynamic}. Additionally, evaluating the quality of T2I-generated images remains an open problem ~\cite{otani2023toward,hartwig2024evaluating}. Commonly used image evaluation metrics, such as CLIP ~\cite{hessel2021clipscore}, Fréchet Inception Distance (FID) ~\cite{heusel2017gans}, and Inception Score (IS) ~\cite{salimans2016improved}, assess different aspects of image quality. However, FID and IS do not evaluate image alignment with the input prompt, making them unsuitable for assessing semantic accuracy in T2I generation. CLIP, on the other hand, measures text-image alignment but treats the image holistically rather than evaluating the quality of individual components, which can result in insignificant differences in CLIP scores between high- and low-quality images. For instance, consider the prompts “a car” versus “a beautiful image of a car parked in front of a big city.” The first prompt is vague and lacks descriptive details, while the second provides richer context. Through our experiments, we observed that images generated from shorter prompts often exhibit structural inaccuracies, such as misaligned or missing components, while longer prompts generally produce more coherent and visually accurate images. However, existing evaluation metrics fail to capture these nuances. CLIP may confirm that both images contain a ‘car,’ but it does not assess whether critical components like tires, doors, windows, or mirrors are correctly positioned and realistically rendered. Without component-aware evaluation, images with structural distortions may still receive high CLIP scores, leading to misleading quality assessments. Addressing these challenges, we propose a component-aware image evaluation framework that automatically refines prompts and evaluates image quality at a component level, enhancing the usability and effectiveness of T2I systems, particularly for users with minimal prompt engineering expertise. In summary, our major contributions are as follows.

\begin{itemize}
    \item We present a fully automated and interactive framework for iterative image generation through user feedback.
    \item We develop a novel image evaluation metric, CAS, to measure the quality of the generated image that addresses the shortcomings of CLIP-based evaluation metrics.
\end{itemize}
\section{PromptIQ Framework}
Our framework consists of five phases as shown in Figure 1.
\begin{figure*}
    \includegraphics[width=\textwidth]{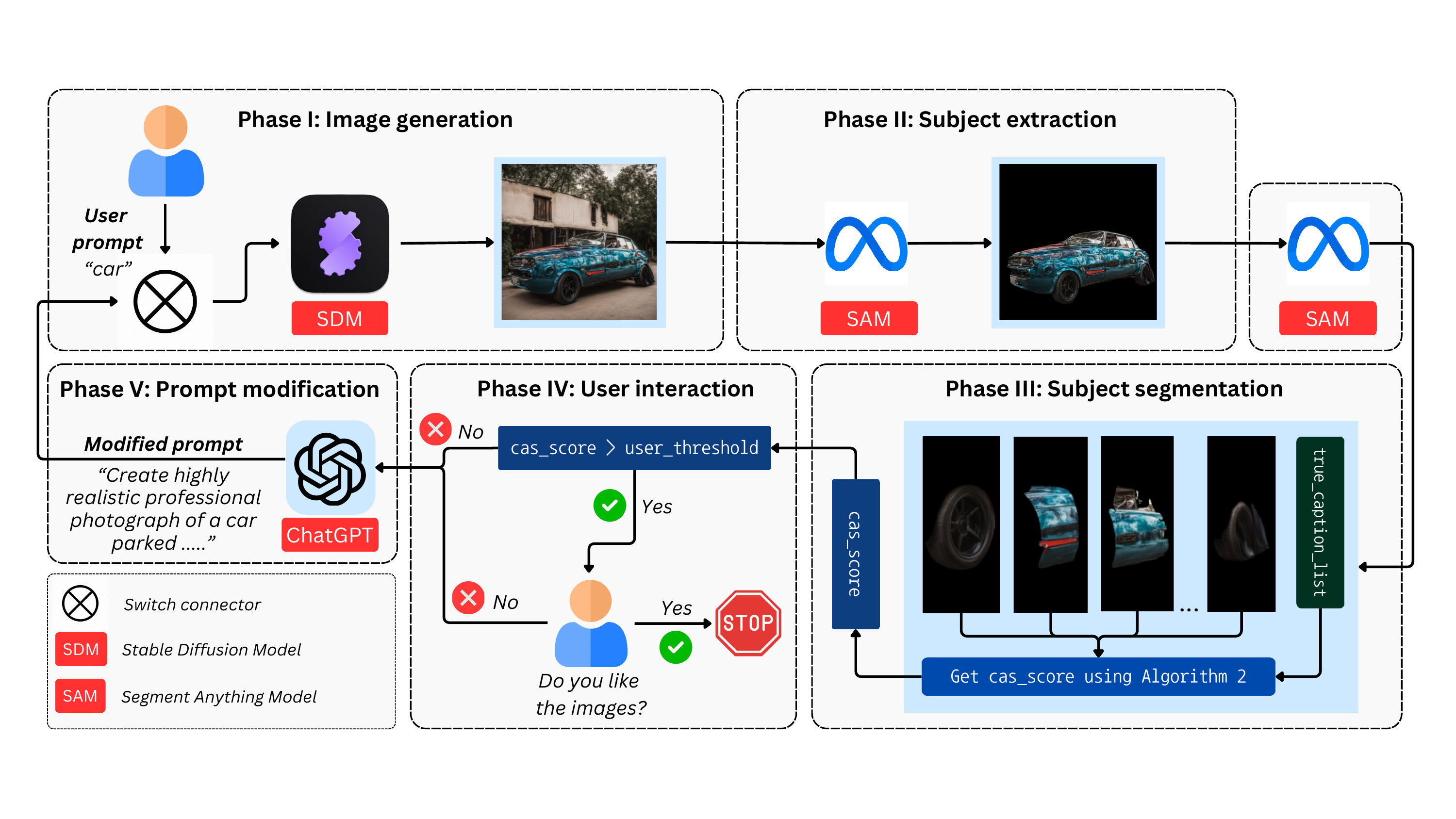}
    \vspace{-13mm}
    \caption{Overview of our automated component-aware framework - PromptIQ, which iteratively refines generated images through a five-phase process, incorporating user feedback to ensure structural accuracy and high-quality results—without the need for prompt engineering.}
\end{figure*}
\newline
\textbf{Phase I: Image Generation.} The image generation process begins when the user gives an initial prompt, which is input for the stable diffusion model (SDM) ~\cite{gupta2024progressive}. Additionally, the user sets {\small \texttt{user\_threshold}}, a parameter that determines the minimum acceptable quality score.
Once the prompt is submitted, SDM generates an image. However, rather than displaying it immediately, we first evaluate its quality to ensure it meets the required standards.
\newline
\textbf{Phase II: Subject extraction.} In this phase, the Segment Anything Model (SAM) ~\cite{kirillov2023segment} eliminates extraneous background elements, effectively isolating the primary subject of the image. This refinement ensures that subsequent evaluations concentrate solely on the subject, minimizing distractions from irrelevant components.
\newline
\textbf{Phase III: Subject Segmentation.} Once the subject is extracted, we segment it further into individual components using SAM. To evaluate the quality of these components, we introduce the {\small\texttt{true\_caption\_list}}, a predefined set of core components essential for defining a subject. For example, for the subject \texttt{\small "car"}, the \texttt{\small true\_caption\_list} includes key components such as:
{\small\texttt{['wheel', 'door', 'headlight', 'mirror', 'windshield', 'grille']}}.
Algorithm \ref{alg:blip_score} is then executed to compute Component-Aware Similarity (CAS) score, {\small\texttt{cas\_score}}, which represents the highest similarity between the generated components and the predefined \texttt{\small true\_caption\_list}. For each segmented component, a BLIP captioning model ~\cite{li2022blip} generates a textual description, which is then compared against all true labels in \texttt{\small true\_caption\_list} using SBERT (SentenceTransformer) ~\cite{reimers2019sentence}. The {\small\texttt{cas\_score}} is updated whenever a component achieves a higher similarity score and is subsequently returned to Algorithm \ref{alg:image_generation} for evaluation.\\
\newline
\textbf{Phase IV: User Interaction.} Once {\small\texttt{cas\_score}} is computed, it is compared to the user-defined threshold, {\small\texttt{user\_threshold}}. If {\small\texttt{cas\_score}} exceeds {\small\texttt{user\_threshold}}, the system displays the generated image to the user and asks for feedback. The user can either accept the image, terminate the process, or reject it, prompting further refinement. 
If {\small\texttt{cas\_score}} is below {\small\texttt{user\_threshold}}, the image is discarded without user exposure, and the process moves to Phase V for prompt modification.
\newline
\textbf{Phase V: Prompt Modification.} The initial prompt undergoes ChatGPT refinement if the generated image does not meet the required quality. The system modifies the prompt to enhance specificity and improve generation quality. The refined prompt is then passed back to the SDM, restarting the process from Phase I. This iterative approach ensures progressive refinement, allowing the system to adapt based on previous generations. 

\begin{figure*}[htbp]
\begin{center}
\includegraphics[width=0.92\textwidth]{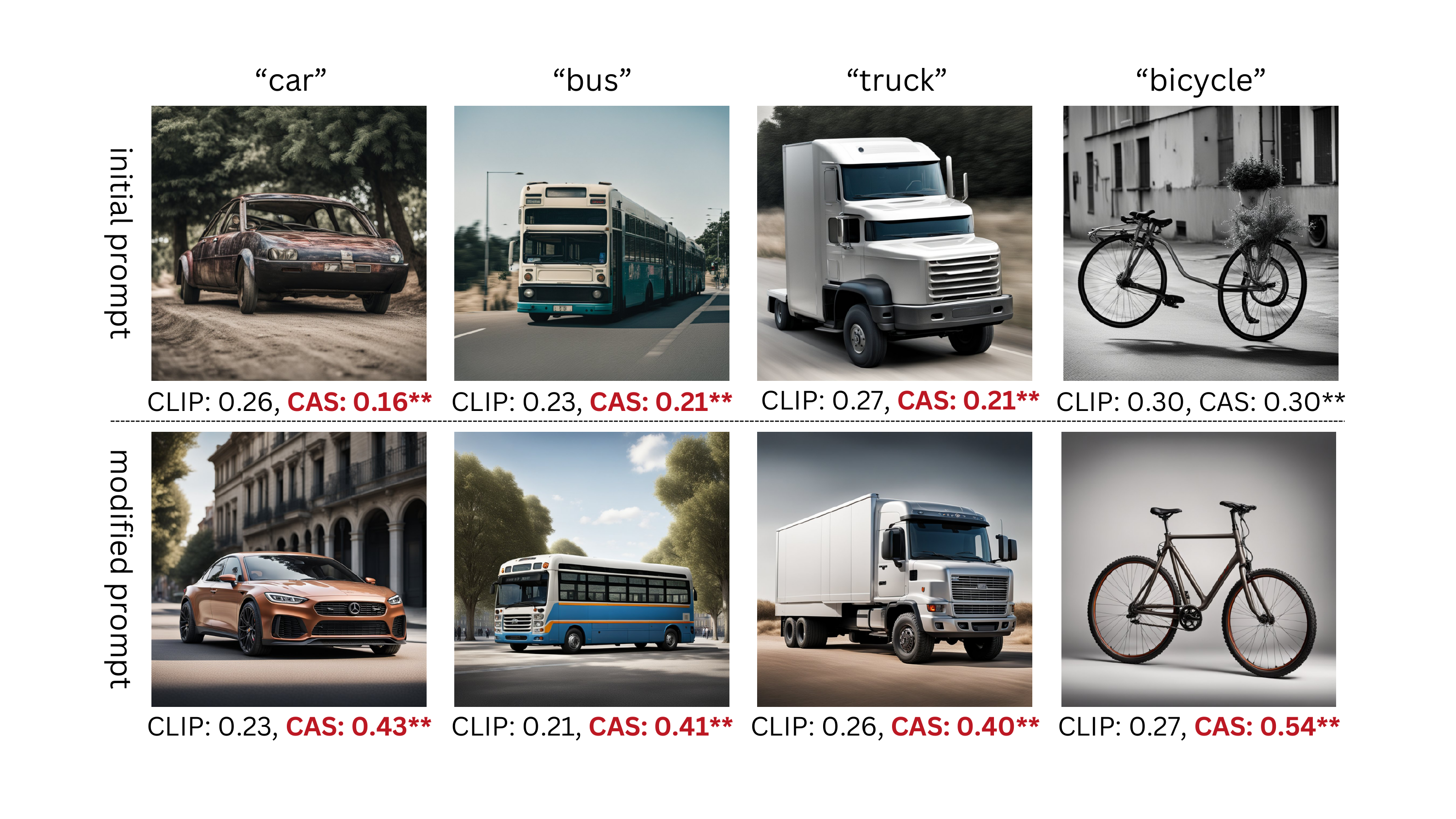}
\vspace{-8mm}  
\caption{ Comparison of CLIP and CAS scores across four subjects—car, bus, truck, and bicycle—before and after prompt modification. CAS effectively differentiates structurally flawed images from well-formed ones, while CLIP remains largely insensitive to these variations.}
\end{center}
\end{figure*}

\begin{algorithm}[!tb]
    { \caption{Interactive image generation and evaluation}
    \label{alg:image_generation}
    \textbf{Input}: $user\_prompt$\\
    \textbf{Parameter}: $user\_threshold = 0.2$\\
    \textbf{Output}: Accepted image or refined prompt

    \begin{algorithmic}[1]
        \STATE Initialize: { $input\_prompt \gets user\_prompt$,\\
        $true\_caption\_list$,$user\_response\_flag$ $\gets False$}
        \WHILE{not $user\_response\_flag$}
            \STATE $image$ $\gets$ \text{SDM($input\_prompt$)}
            \STATE Extract $subject$ using SAM ($image$)
            \STATE Get $component\_masks$ using \text{SAM($subject$)}
            \STATE Compute $cas\_score$ using Algorithm \ref{alg:blip_score}
            \IF{$cas\_score > user\_threshold$}
                \STATE Display \texttt{image} to user
                \STATE {Ask user: {\small \texttt{"Do you like the image? 'yes' or 'no'"}} $\rightarrow user\_response$}
                \IF{$user\_response = \texttt{"yes"}$}
                    \STATE Set $user\_response\_flag \gets True$
                    \STATE \text{STOP!}
                \ELSE
                    \STATE Update $input\_prompt$ using \text{ChatGPT} 
                    \STATE Go to Step 2
                \ENDIF
            \ENDIF
        \ENDWHILE
    \end{algorithmic}}
\end{algorithm}

\begin{algorithm}[tb]
    \caption{Compute component-aware similarity score}
    \label{alg:blip_score}
    \textbf{Input}: $component\_masks$, $true\_caption\_list$\\
    \textbf{Output}: component-aware similarity score $cas\_score$

    \begin{algorithmic}[1]
        \STATE Initialize $cas\_score \gets 0$
        \FOR{each valid mask in $component\_masks$}
            \STATE $extracted\_component \gets $ Extract segmented region
            \STATE $g\_caption \gets \text{BLIP($extracted\_component$)}$
            \FOR{ $true\_label$ in $true\_caption\_list$}
                \STATE $comp\_score \gets \text{SBERT($g\_caption$, $true\_label$})$ 
                \STATE Update $cas\_score$ if greater than $comp\_score$
            \ENDFOR
        \ENDFOR
        \STATE \textbf{return} $cas\_score$
    \end{algorithmic}
\end{algorithm}

\section{Experiments}  

\subsection{Experimental Setup}   
We implemented the proposed framework in Python and deployed it with Gradio, enabling an interactive user interface for real-time image generation and evaluation. All experiments and evaluations were conducted using an NVIDIA A100 GPU on Google Colab.

\subsection{Results and Analysis}  
The evaluation was conducted on four subjects: car, bus, truck, and bicycle to assess the effectiveness of our CAS compared to CLIP using initial prompts - extremely short and ambiguous prompts (e.g., “car”) and modified prompts generated by ChatGPT. Images from initial prompts were often structurally flawed, with missing or distorted components, as the T2I model struggled with ambiguous inputs. In contrast, modified prompts produced well-formed images, demonstrating the importance of structured guidance in a generation.

CLIP scores failed to differentiate image quality, remaining similar across flawed and refined images (e.g., 0.23–0.30 for initial prompts, 0.21–0.27 for modified prompts). In contrast, CAS accurately distinguished between poor and high-quality generations, assigning low scores to images with missing or misaligned components (e.g., 0.16) and higher scores to well-structured outputs (up to 0.54). This demonstrates that CAS effectively captures structural coherence, unlike CLIP.

\section{Conclusion and Future work}
In this work, we introduce an automated, component-aware evaluation framework designed to iteratively refine generated images until users are satisfied, removing the need for prompt engineering expertise. Unlike CLIP, which overlooks structural inconsistencies, our CAS score provides a more accurate and human-aligned assessment, effectively distinguishing between flawed and well-structured images. This capability is particularly crucial in fields where structural precision matters, such as medical imaging, autonomous driving, and industrial design. In the future, we plan to expand user control by incorporating multiple diffusion models and user-driven prompt refinement, allowing for even greater flexibility and higher-quality outputs.

\bibliographystyle{named}
\bibliography{promptIQ}

\end{document}